\documentclass{vgtc}                          

\ifpdf
  \pdfoutput=1\relax                   
  \usepackage{graphicx}                
  \DeclareGraphicsExtensions{.pdf,.png,.jpg,.jpeg} 
\else
  \usepackage{graphicx}                
  \DeclareGraphicsExtensions{.eps}     
\fi%

\graphicspath{{figures/}{pictures/}{images/}{./}} 

\usepackage{microtype}                 
\PassOptionsToPackage{warn}{textcomp}  
\usepackage{textcomp}                  
\usepackage{mathptmx}                  
\usepackage{amsmath}
\usepackage{times}                     
\usepackage{cite}                      
\usepackage{tabu}                      
\usepackage{booktabs}                  

\vgtccategory{Research}

\vgtcinsertpkg

\title{Crop Planning using Stochastic Visual Optimization}

\author{Gunjan Sehgal\thanks{e-mail: sehgal.gunjan@tcs.com}, Bindu Gupta\thanks{e-mail:bindu.gupta2@tcs.com}, Kaushal Paneri, Karamjit Singh, Geetika Sharma, Gautam Shroff\thanks{e-mail:gautam.shroff@tcs.com}}
\affiliation{\scriptsize TCS Research, India} 

\abstract{ As the world population increases and arable land decreases, it becomes vital to improve the productivity of the agricultural land available. Given the weather and soil properties, farmers need to take critical decisions such as which seed variety to plant and in what proportion, in order to maximize productivity. These decisions are irreversible and any unusual behavior of external factors, such as weather, can have catastrophic impact on the productivity of crop. A variety which is highly desirable to a farmer might be unavailable or in short supply, therefore, it is very critical to evaluate which variety or varieties are more likely to be chosen by farmers from a growing region in order to meet demand. In this paper, we present our visual analytics tool, \textit{ViSeed},  showcased on the data given in Syngenta 2016 crop data challenge\footnote{https://www.ideaconnection.com/syngenta-crop-challenge/}. This tool helps to predict optimal soybean seed variety or mix of varieties in appropriate proportions which is more likely to be chosen by farmers from a growing region. It also allows to analyse solutions generated from our approach and helps in the decision making process by providing insightful visualizations.} 

\CCScatlist{ 
  \CCScat{K.6.1}{Management of Computing and Information Systems}%
{Project and People Management}{Life Cycle};
  \CCScat{K.7.m}{The Computing Profession}{Miscellaneous}{Ethics}
}

\begin{document}


\firstsection{Introduction}

\maketitle

\par
With increasing world population and decreasing arable land, optimizing the productivity of land is the need of the hour.
To meet the growing food demand~\cite{godfray2010food} farmers need to take critical decisions like \textit{on what}(soil), \textit{which} (variety), and \textit{how much}(proportion) in order to maximize the yield. However, a seed variety highly desired by a farmer may be unavailable or in short supply. Therefore, it is critical to evaluate which variety is more likely to be chosen by farmers from a growing
region in order to meet demand. As the production varies due to differences in weather conditions~\cite{iizumi2015weather} and soil quality~\cite{drummondstatistical}, there cannot be a single seed variety that will yield high production across regions. Hence, predicting the appropriate mix of proportions of different seed varieties is desirable. 

While a standalone solution with predicted seed variety or mix of varieties for a region is very helpful, but in the real world, where the number of regions and varieties both can be large, it is very crucial to have visualization based platform to analyze all possible solutions. In this paper, we present a machine learning and optimization based approach to predict seed variety or mix of varieties in appropriate proportions which is more likely to be chosen by farmers from a growing region. We also present our visual analytics tool to understand and analyse the solutions generated by our approach. This tool helps in the decision making process and also provides insightful visualizations.

We use Syngenta Crop Challenge 2016 dataset which requires computation of an optimal combination of upto five soybean varieties for the given region using historical sub-region data and experiment data from a few experiment locations. Our approach to the challenge comprises of (a) machine learning and optimization techniques to compute the solution, as well as (b) a geo-spatial visual analytics tool, ViSeed, to understand the given data and analyse the solutions.
\par
As the production of soybean varies due to differences in weather conditions and soil quality and there is no single variety that will yield high production across the region, we have experimented with two possible solutions to the challenge. The first, global solution, required by the challenge and defined as a combination of upto five varieties to be grown over the entire region. The second, a differentiated solution, consisting of individual optimal combinations for each sub-region. We also compute a spatial cohesion score for each subregion which measures the similarity of the solutions for the sub-region with those of its neighbors. Finally, we show that the differentiated solution performs marginally better and has lower variance than the global solution.

Rest of the paper is organized as follows. In section~\ref{theory}, we present the problem statement along with out approach to predict the optimal set of varieties and in section~\ref{visual}, we present our visual analytics tool ViSeed and showcase it for the Syngenta data challenge 2016. In section~\ref{exp}, we present our results of prediction accuracies and predicted soyabean varieties and conclude in section~\ref{con}.



\section{Methodology and theory	}\label{theory}
\subsection{Problem Statement:} Let $\{v_1, v_2,..., v_n\}$ be the $n$ soybean varieties and let $R$ be a region with $k$ sub-regions $\{R_1, R_2,..., R_k\}$. Each of the sub-region $R_i$ has different soil and weather conditions associated with it. Our goal is to find which soybean seed variety, or mix of up to five varieties in appropriate proportions, will best meet the demands of farmers in each sub-region $R_i$ and for whole region $R$. In order to find such mix of seed varieties, we predict the top five varieties for each sub-region in terms of maximum yield and then find the optimal proportion of these five varieties in order to maximize yield and minimize variance in the yield. 

Our approach to predict the mix of soybean varieties in appropriate proportion for each subregion $R_i$ and for whole region $R$ consists of three steps: 1) Prediction of Weather and Soil attributes, 2) Yield Prediction given the soybean variety, and weather and soil conditions, and 3) Yield Optimization. We also present our visual analytic tool to understand and analyses the solutions generated by our approach. This tool helps in decision making process and also provide insightful visualizations.

\subsection{Weather and Soil Prediction} 
\begin{figure}
\centering
\includegraphics[width = 80mm]{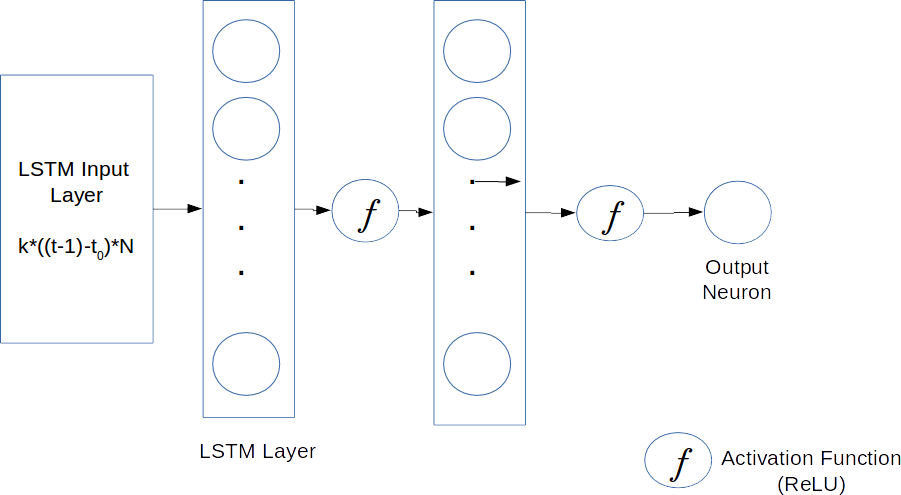}
\caption{LSTM architecture used to predict weather and soil attributes}
\label{fig:lstm}
\end{figure}

For each subregion $R_i$, let $\{w^i_1(t), w^i_2(t),..., w^i_N(t)\}$ be the weather condition and $\{s^i_1(t), s^i_2(t),..., s^i_M(t)\}$ be the soil condition attributes at time $t$. Given these attributes from initial time $t_0$ to the current time $t$, we use Deep learning based approach \textit{LSTM}~\cite{hochreiter1997long} to predict the value of these attributes for time $t+1$. For each attribute separately, say $w^i_N$, we prepare a sequence of its values from time $t_0$ to $t-1$, and use it as an input to train LSTM with the target of predicting value of $w^i_N$ at time $t$ and then we use the trained LSTM to predict the value for time $t+1$.LSTM based neural networks are competitive with the traditional methods and are considered a good alternative to forecast general weather conditions~\cite{articleLSTM} . Figure~\ref{fig:lstm} shows an architecture used for the prediction of weather and soil attributes. 

\begin{figure*}
\begin{center}
\begin{tabular}{cc}
\includegraphics[height=1.8in]{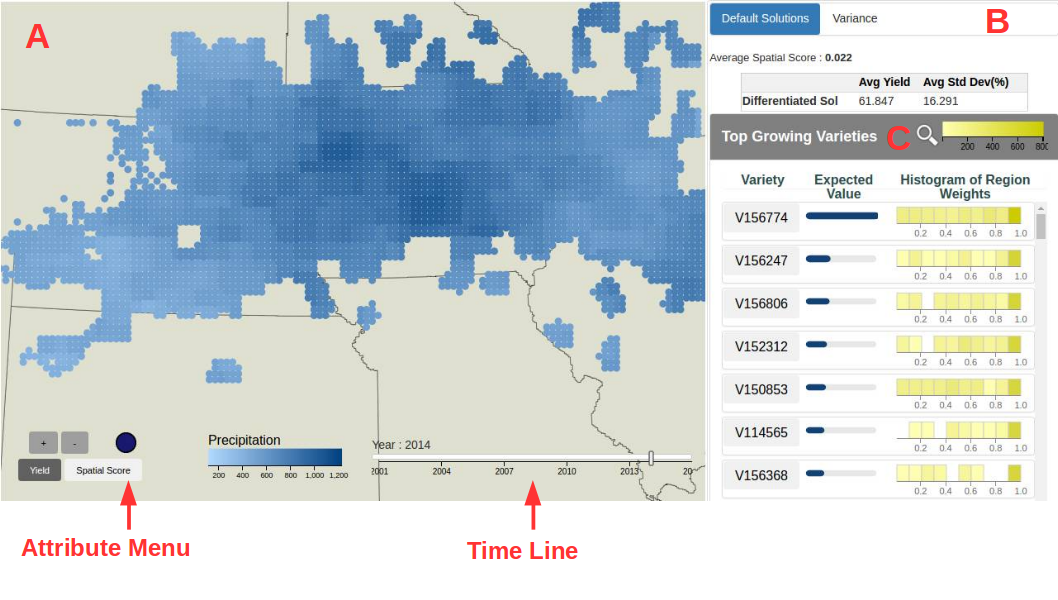} & \includegraphics[height=1.8in]{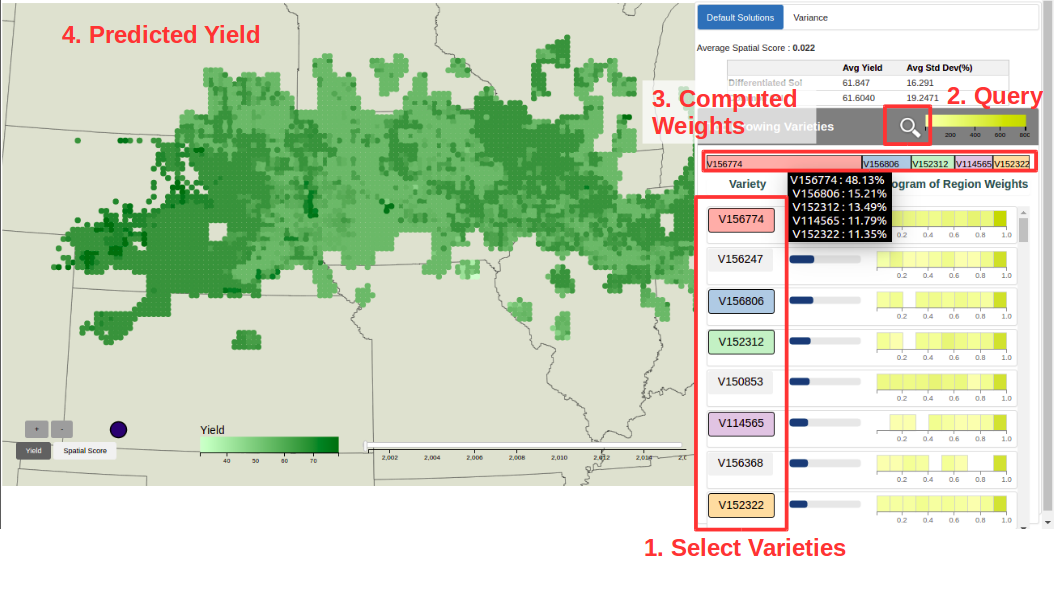}\\
(i) & (ii)
\end{tabular}
\end{center}
\caption{ViSeed: (i) Main Data Exploration Screen (ii) Common Solution Exploration} \label{MainScrn}
\end{figure*} 
 
\subsection{Yield Prediction}
Once we predict the weather and soil attributes for the time $t+1$, we use these attributes as a feature set to predict the yield in every sub-region $R_i$ for each soybean variety $v_j$, where $j =1,2,..., n$. We divide the yield value into $r$ equally sized bins by taking maximum and minimum from historical data and we treat the prediction problem as a classification problem,  where our goal is to predict the bin value of yield. More formally, for each sub-region $R_i$, we compute $n$ probability distributions $\{p^i_{j,1}, p^i_{j,2},..., p^i_{j,r}\}$ of yield, one for every soybean variety $v_j$, where $j = 1,2,..., n$, using Random Forest Classifier (RFC)~\cite{breiman2001random}. RFC is an ensemble learning method that operates by constructing a multitude of decision trees at training time and outputting the class that is the mode of the classes (classification) of the individual trees. As an output, we get the count of each class which represents the number of trees outputting the class. Further, we convert these counts into probabilities by diving each count by the sum of all counts.
\subsection{Yield Optimization}
Given $n$ probability distributions of the yield $y_i$ for sub-region $R_i$, we use an optimization approach to obtain weighted combination of varieties in order to maximize yield and minimize standard variation or variability. Steps to obtain combination of varieties in an appropriate proportion are given as follows:

\begin{enumerate}

\item Given n probability distributions $\{p^i_{j,1}, p^i_{j,2},..., p^i_{j,r}\}$ of yield $y_i$ for every soybean variety $v_j$, where $j = 1,2,..., n$, we calculate expected value and variance of each distribution, represented by $E_j$, and $Var_j$ respectively.

 \item We choose top $k$ distributions out of $n$ having maximum score calculated as:
 \begin{equation}
  score_j = norm(E_j) + (1 - norm(Var_j))
 \end{equation}
where $norm(E_j)$ and $norm(Var_j)$ are the normalized values between $0$ to $1$ of $E_j$ and $Var_j$ respectively.

\item We use an optimization technique with objective function of maximizing yield using combination of upto five varities out of chosen top $k$ in an appropriate proportion. The objective function and constraints of optimization are given as follows:
\begin{equation}
\textrm{Objective function:  } max (\sum_{l}w_l*E_l)
\end{equation}
\begin{equation}
 \textrm{Constraints:   } \sum_{l}{w_l*norm(Var_l)} < \tau 
 \textrm{ , } w_l > .10
  \textrm{, and } \sum_{l} w_l = 1
\end{equation}
 
The term $\sum_{l}{w_l*norm(Var_l)}$ is called \textit{variability}. We ran above optimization for ten thresholds of variability, i.e., ten values of $\tau$ from $0$ to $1$ with the step size of $0.1$. For every value of $\tau$, we get the optimal solution. Therefore, for a sub-region $R_i$, we get 10 solutions and out of these 10 solutions, we choose the solution having maximum yield and minimum variability and it is called \textit{default solution} for sub-region $R_i$
\end{enumerate}

\textbf{Spatial Cohesion(SC) Score:} For every solution obtained from optimization approach, we calculate SC Score which is calculated as follows: Let for a sub-region $R_i$, the optimized solution at $\tau = \tau_1$ contains five varieties $v^i_1, v^i_2,..., v^i_5$ in some proportion. Let $near(R_i) = \{ N^i_1, N^i_2,..., N^i_{l_i}\}$ be the set of neighboring sub-regions of $R_i$ which are having maximum distance of $m$ miles from the centroid of it. For every $v^i_j$ in the solution of $R_i$, we calculate variety score as:
\begin{equation}
 var_{s}(v^i_j) = \dfrac{\sum^{k = l_i}_{k=1}{w^i_{j,k}}}{l_i}
\end{equation}
where $w^i_{j,k}$ is the proportion with which variety $v^i_j$ exist in the solution of neighboring sub-region $N^i_k$ at $\tau = \tau_1$. Further, SC score of sub-region $R_i$ is calculated as the average of all variety scores in the solution:
\begin{equation}
 SC_i = \dfrac{\sum^5_{j =1}{var_{s}(v^i_j)}}{5}
\end{equation}

\section{Visual analytics using ViSeed}\label{visual}

\par
 We now describe, ViSeed, our visual analytics tool to understand the given data and analyse the solution generated by our analytics methodology.This tool differentiates our work from other related works~\cite{adekanmbi2015multiobjective}~\cite{marko2016soybean} as it lets the retailer explore varieties performing well in local as well as global areas. As agricultural yields vary widely around the world due to climate and the mix of crops grown~\cite{foley2011solutions},ViSeed lets the farmer explore the quality of soil and climatic variations region wise intuitively to take the planting decisions accordingly.
\par
The main screen of ViSeed is divided into two parts. The first part displays a map of the United States over which various sub-region attributes can be visualized, Figure \ref{MainScrn} (i) A. The second (right hand panel) part, B, contains a tabbed control panel, to switch between various visualizations of the solution data. 

\par
\textbf{Getting Started} A data attribute such as precipitation or solar radiation may be visualized by selecting it from the attribute menu and a year from the timeline as shown in Figure \ref{MainScrn} (i). Further, as described in the previous section, we first compute the top $k$ varieties for each sub-region, based on high expected yield and low variance and from them, an optimal solution with up to five varieties. As a variety may occur in the optimal solution of multiple sub-regions, we compute a distribution of weights for each variety across sub-regions and the expected value of this distribution. The varieties are ranked in decreasing order of expected value which is an indicator of prevalence of the variety across sub-regions and displayed in the right panel when the visualization tool starts up, Figure \ref{MainScrn} (i),  C. Thus, the user is provided a starting point to begin exploring the possible solutions.
\par
The histogram of weights for each variety across sub-regions, using a colour map, is shown alongside its expected value. Selection of a range on the histogram of a variety highlights those sub-regions for which the variety has weight or proportion in the selected range. Clicking on a variety name highlights those sub-regions on the map for which the variety occurs in the optimal solution. Clicking multiple varieties highlights the regions for all, thus allowing the user to visualize cumulative prevalence of varieties. 
\par
\textbf{Common Solution} A user may explore various solutions, for the entire region, by selecting up to five varieties from this list,  1 in Figure \ref{MainScrn} (ii). On pressing the query button, 2, proportions of each of the selected varieties are computed, 3, and the total yield for the region is predicted, 4. 
\par
\begin{figure}[tbp]
\centering
\begin{minipage}[b]{0.43\textwidth}
\includegraphics[height=1.8in]{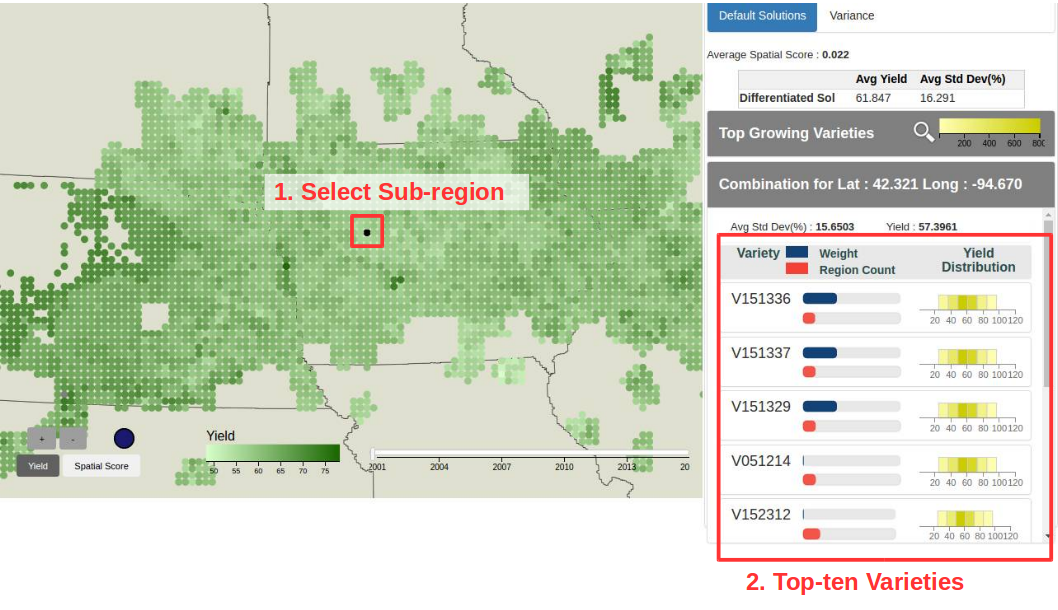}
\caption{Top $k$ Varieties for a Sub-Region} \label{TopTen}
\end{minipage}
\hspace{1cm}
\begin{minipage}[b]{0.43\textwidth}
\includegraphics[height=1.8in]{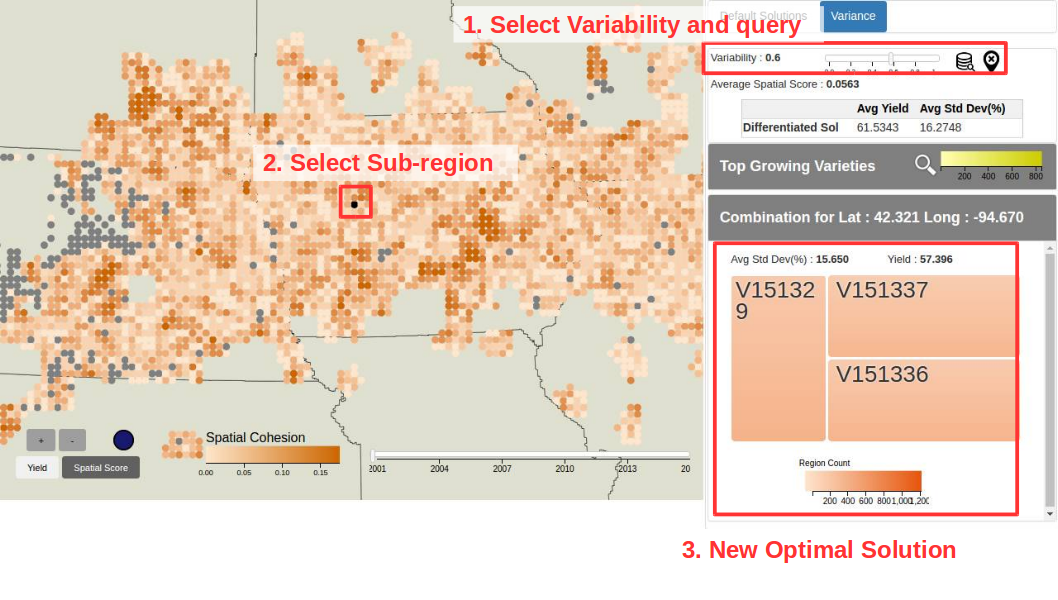}
\caption{Changing Solution Variability} \label{Variance}
\end{minipage}
\end{figure}

\textbf{Differentiated Solution} Each sub-region has a precomputed default solution from among its top $k$ varieties. Clicking on a sub-region in the map, Figure \ref{TopTen}, 1, brings up the top $k$ varieties for that sub-region, 2, along with their weights in the optimal solution, count of sub-regions in which the variety is in the top $k$ and the predicted yield distribution for the variety. The sub-regions for which a variety is in the top $k$ list can be seen by clicking on the red count bar. 

For every solution, we show the \textit{Average Yield} and \textit{Average Standard Deviation} and \textit{Average offset in \%} of entire region $R$. The standard deviation and offset of each sub-region is calculated as:
\begin{equation} 
 \textrm{Standard deviation (S.D)} = \dfrac{\sum^5_{l=1}{w_l*\sqrt{Var_l}}}{5} 
\end{equation}
\begin{equation}
\textrm{Offset \%} = (\dfrac{S.D}{Expected Yield})*100 
\end{equation}

where $Var_l$ is the variance of variety $v_l$ and $w_l$ is its proportion in the solution.  

\par
\textbf{Changing Variability} Our tool also allows the user to analyse solutions by playing with different variability thresholds. This can be done by first clicking on the Variance tab. The user can set a variability threshold by moving the slider, Figure \ref{Variance}, 1. On pressing the query button, the list of varieties, with variability below the chosen threshold is displayed, along with their histograms of weights across sub-regions and expected values as before. A new optimal solution is computed for each sub-region and its similarity with the solutions of neighboring sub-regions is calculated as a spatial cohesion score. This score is visualized on the map. The user may also interact with this list of varieties and compute a global solution for the entire region, as described earlier. 

\section{Quantitative results}\label{exp}
In this section, we present our results of LSTM and RFC model used for weather and yield prediction. We used keras, scikit-learn and cvxpy respectively to implement LSTM, RFC and optimization.

\textbf{Available Data:} We were provided with the following two datasets:

\begin{enumerate}
 \item \textbf{Experiment Dataset:} It consists of 82000 experiments, in 583 sub-regions, between 2009 and 2015 using 174 varieties. It has three weather and three soil condition attributes for every sub-region.
 \item \textbf{Region Dataset:} It consists of 6490 sub-regions with the given three soil and three weather condition attributes from the year 2000 to 2015. 
\end{enumerate}

We predict three attributes of weather conditions, temperature, precipitation, and solr radiation for every sub-region of Region dataset. For an attributes say $w^i_N$ in sub-region $R_i$, we prepare the sequence of 14 values from year 2000 to 2014 as an input to train LSTM with a target of predicting value of year 2015. We validate and test the LSTM model by dividing Region dataset into training, validation, and test data in the ratio of 70:15:15 respectively. Table~\ref{table:lstm}, shows the \textit{normalized root mean square error (N-RMSE)} for all three weather attributes on validation and test set. Here, N-RMSE is defined for an attribute say, $w^i_N$ as 
\begin{equation}
 \textrm{RMSE} = \sqrt{\dfrac{\sum^{k}_{i=1}{(\textrm{act}(w^i_N) - \textrm{pred}(w^i_N))}^2}{k}}
\end{equation}
\begin{equation}
 \textrm{N-RMSE } = \dfrac{RMSE}{(max(w^i_N)-min(w^i_N))} * 100
\end{equation}
where $act(w^i_N)$ and $pred(w^i)$ are the actual and predicted values, $max(w^i_N)$ and $min(w^i)$ are maximum and minimum values, of attribute $w^i_N$ for sub-region $R_i$. The value of $k$ is $6490$ in our case.
\begin{figure}[tbp]
\begin{center}
\begin{tabular}{c}
\includegraphics[height=1.75in]{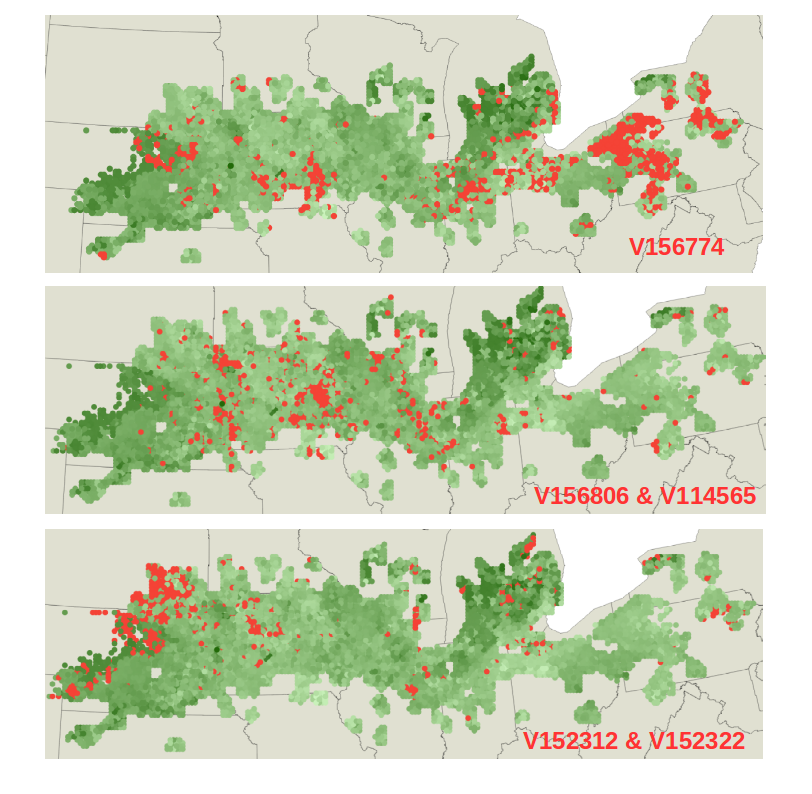} \\
(i) \\
  \includegraphics[height=1.75in]{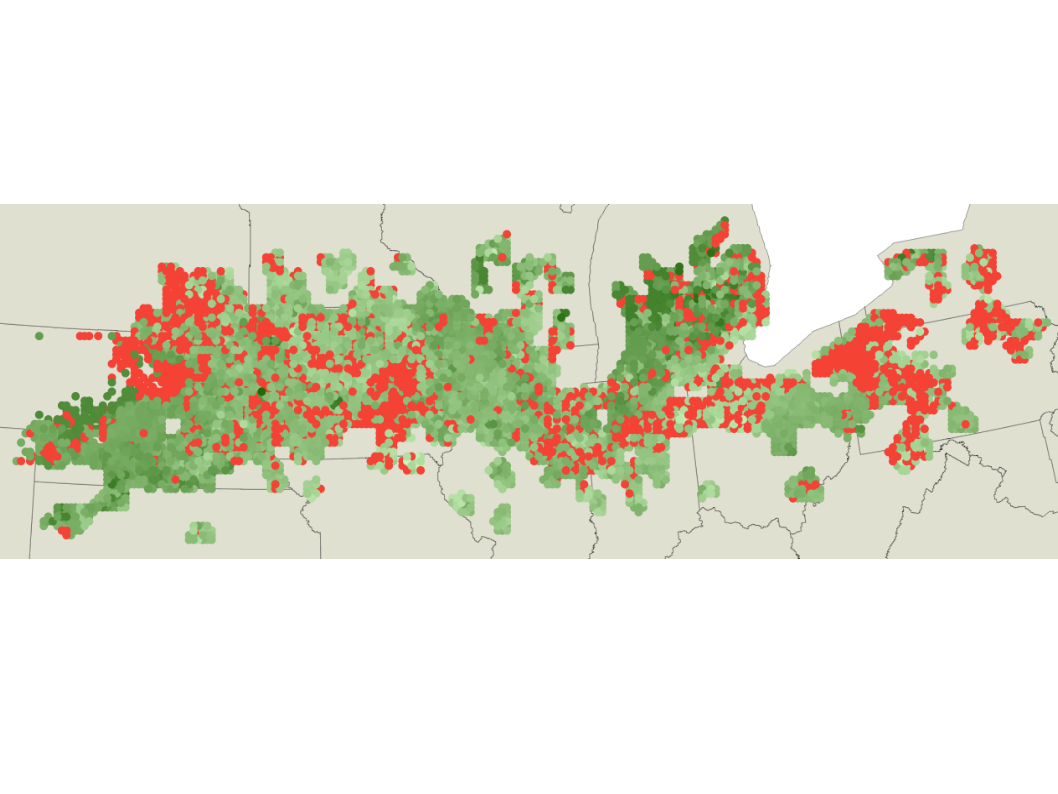}\\
  (ii)
\end{tabular}
\end{center}
\caption{Common Solution: (i) Areas in which Varieties are optimal (ii) Cumulative area in which varieties are optimal} \label{Regions}
\end{figure}
N-RMSE in Table~\ref{table:lstm} indicating that the prediction of weather attributes using LSTM have less than 1\% error for temperature and precipitation and less than 3\% error for Solr radiation. We use this trained LSTM model to predict weather attributes for year 2016. Note, we did not predict soil condition parameters as they did not change over time in experiment dataset.

We use experiment dataset to train RFC model by dividing it into three parts train, valid and test dataset in the ratio of 70:15:15 respectively. Soil and weather condition attributes in each experiment has been used as a feature set in RFC and discretized yield as a target variable. N-RMSE of yield predicted on validation and test set using RFC is \textbf{6.01\%} and \textbf{6.25\%} respectively. We use the trained RFC model to predict the yield for 6490 sub-regions in Region dataset with weather and soil attributes for year 2016 as input, predicted using LSTM(as explained above).

\renewcommand{\arraystretch}{}
\begin{table}
  \centering
  \begin{tabular}{|c|c|c|c|} 
   \hline
   Attributes & Validation Set & Test Set \\
   \hline
   $Temperature$ & 0.69\% & 0.78\% \\
   \hline
   $Percipitation$ & 0.73\% & 0.83\% \\
   \hline
   $Solr Radiation$ & 2.6\% & 2.8\%  \\
   \hline 
  \end{tabular}
  \vspace{6pt}
\caption{N-RMSE of three weather attributes on validation and test set using LSTM}
  \label{table:lstm}
\end{table}

\subsection{Insights from ViSeed}

\textbf{Common Solution} In order to compute the common solution, we used ViSeed to check multiple combinations of the top ten growing varieties based on expected value and arrived at the reported solution. The areas in which these varieties are in the optimal differentiated solution are shown in figure \ref{Regions} (i), and the cumulative area for all five is shown in figure \ref{Regions} (ii).

\par
\begin{figure*}
\begin{center}
\begin{tabular}{cc}
\includegraphics[height=1.75in]{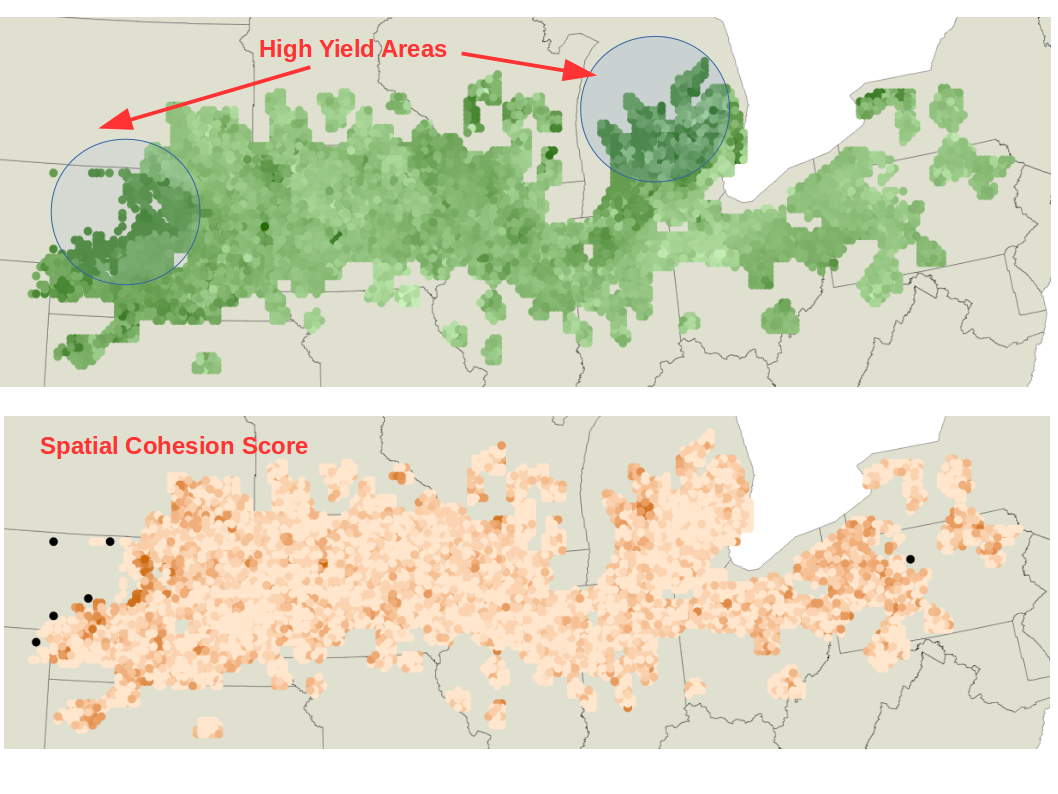} & \includegraphics[height=1.75in]{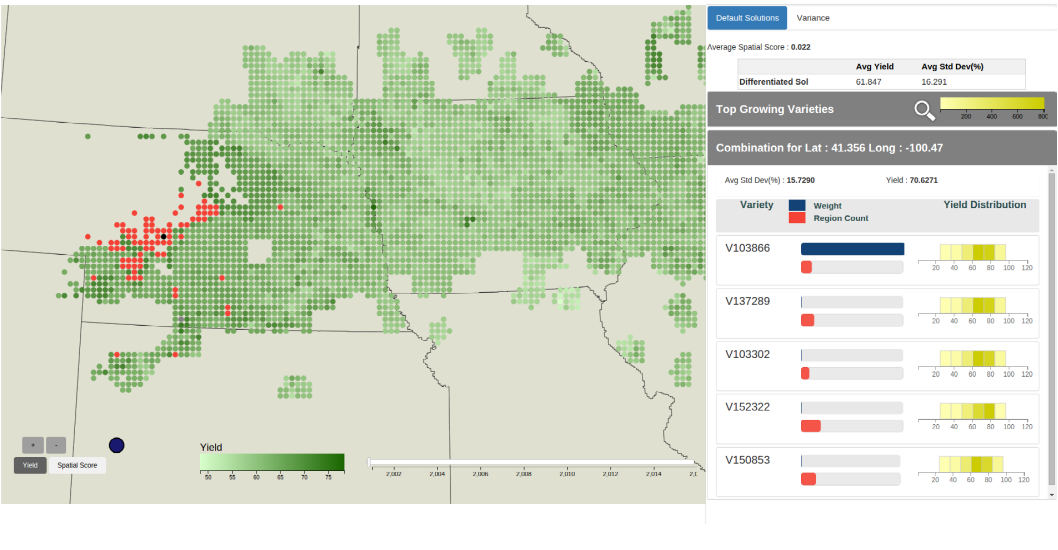}\\ 
 (i) & (ii)
\end{tabular}
\end{center}
\caption{Differentiated Solution: (i) High Yield Areas and Spatial Cohesion (ii) Validation of Spatial Cohesion} \label{DiffSoln}
\end{figure*}

\textbf{Differentiated Solution} Analysing the predicted yield for the differentiated solution, we find two areas with high yield as highlighted in figure \ref{DiffSoln} (i). The spatial cohesion score is also visualised in the same figure and observers to be high for the areas with high yield. We validate this in figure \ref{DiffSoln} (ii), by visualizing the sub-regions in which a variety from a high yield, high spatial cohesion score sub-region is grown.  We may conclude that high yield varieties are localized to certain regions, and so they do not occur in the optimal common solution.  

\begin{figure}[t]
\begin{center}
\includegraphics[height=2.0in]{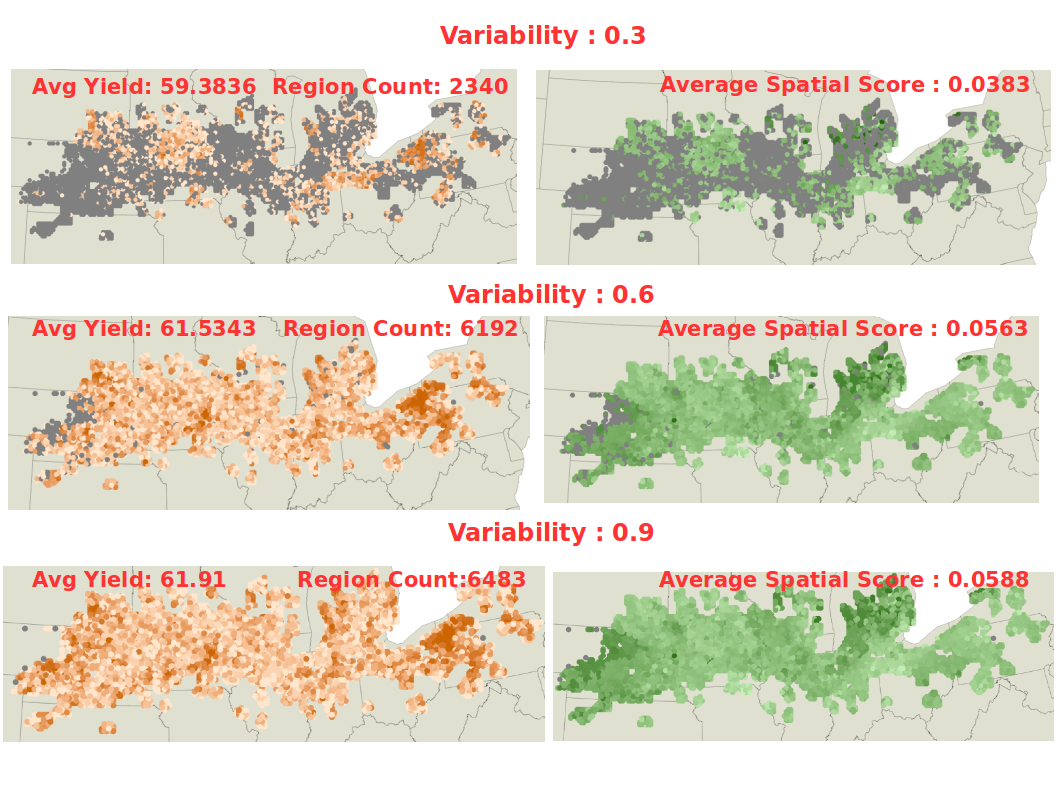}
\end{center}
\caption{Differentiated solutions at different Variability thresholds} \label{Variability}
\end{figure}
\par
\textbf{Experiments with Variability} 
In figure \ref{Variability}, we show the changes in spatial cohesion and yield for different variability thresholds. We find that with a low variability threshold of 0.3, there is no optimal solution for most of the sub-regions (dark-grey areas in the map). Increasing the threshold, results in solutions for most of the sub-regions along with an increase in the average yield per sub-region and the spatial cohesion score. However, beyond a certain threshold, the gain in yield and spatial cohesion is very small.  

Our submission to the Syngenta challenge comprised of two parts. We use (a) machine learning and stochastic optimisation to compute solutions and (b) have developed a visual analytics tool, ViSeed, to analyse the results. In particular, we predict weather and soil attributes of each sub-region through time-series regression using LSTMs. For each sub-region and seed variety, a Random Forest classifier is trained on experiment data to predict yield distributions. Next, we compute weights of varieties through stochastic optimization, maximizing expected yield and minimizing variance, for each subregion, followed by visual analytics to choose an optimal global solution based on the spread of each variety across locally optimal combinatios.. Our combination of soybean varieties is as follows - (i) V156774: 48.1\%, (ii) V156806: 15.2\%, (iii) V152312: 13.5\%, (iv) V114565: 11.8\% and (v) V152322: 11.4\%. We provide a second solution to the challenge consisting of individual optimal solutions for each sub-region which performs marginally better than the first reported above. Our geo-spatial visual analytics tool ViSeed, is designed to explore the raw data as well as aid in optimization. Our entry was not among the top 5 final entries selected from 600 registered teams and as the details of the winning entries have not been made public, we cannot compare our approach with theirs. 

\section{Conclusion}\label{con}
In this paper, we propose an approach for crop planning based on machine learning models (RFC and LSTM), stochastic optimization and a visual analytics platform. We have given 2 different solution sets; i) Common solution for entire region, ii) Differentiated solutions at sub-region level. We use expected yield based on a model using sub-region wise predictions of weather and soil conditions, standard deviation of expected yield as the criteria to select seed varieties. We give the spatial cohesion score for each solution which helps to find the similar solutions in the neigbouring sub-regions. We also present a geo-spatial visual analytics tool which has the capability of exploring raw data and helps the retailer in decision making by allowing exploration of solutions at sub-region as well as global level. 

\bibliographystyle{abbrv}

\bibliography{SVO}
\end{document}